 \documentclass[smallabstract,smallcaptions]{dccpaper}

\usepackage{epsfig}
\usepackage{amsmath}
\usepackage{caption}
\usepackage{amssymb}
\usepackage{color}
\usepackage{graphicx}
\usepackage{siunitx}
\usepackage{floatrow}
\usepackage{booktabs}
\usepackage{subcaption} 
\usepackage{hyperref}
\usepackage{url}

\newlength{\figurewidth}
\newlength{\smallfigurewidth}
\DeclareMathOperator{\mse}{MSE}
\DeclareMathOperator{\ce}{CE}

\begin{document}

\title
{
\textbf{Domain Adaptation for Learned Image Compression with Supervised Adapters}
}

\author{%
Alberto Presta$^{\ast}$, Gabriele Spadaro$^{\ast}$, Enzo Tartaglione$^{\dag}$, \\ Attilio Fiandrotti$^{\ast \dag}$ and Marco Grangetto$^{\ast}$\\[0.5em]
{\small\begin{minipage}{\linewidth}\begin{center}
\begin{tabular}{ccc}
$^{\ast}$University of Turin & \hspace*{0.5in} & $^{\dag}$LTCI, Télécom Paris, \\
Computer science department && Institut Polytechnique de Paris \\
 Turin, Italy && Paris, France\\
\texttt{\{name.surname\}@unito.it} && \texttt{\{name.surname\}@telecom-paris.fr}
\end{tabular}
\end{center}\end{minipage}}
}

\maketitle
\thispagestyle{empty}

\begin{abstract}
In Learned Image Compression (LIC), a model is trained at encoding and decoding images sampled from a source domain, often outperforming traditional codecs on natural images;
yet its performance may be far from optimal on images sampled from different domains.
In this work, we tackle the problem of adapting a pre-trained model to multiple target domains by plugging into the decoder an adapter module for each of them, including the source one. 
Each adapter improves the decoder performance on a specific domain, without the model forgetting about the images seen at training time.
A \emph{gate network} computes the weights to optimally blend the contributions from the adapters when the bitstream is decoded.
We experimentally validate our method over two state-of-the-art pre-trained models, observing improved rate-distortion efficiency on the target domains without penalties on the source domain.
Furthermore, the gate's ability to find similarities with the learned target domains enables better encoding efficiency also for images outside them.
\end{abstract}

\section{Introduction}\label{intro}

Learned image compression (LIC)  has gained considerable interest as it may achieve equal or even better compression efficiency than standardized codecs~\cite{review}.
In this context, a parametric autoencoder is trained to project an image into a latent representation that is quantized and entropy-coded as a compressed bitstream.
On the decoder side, the latent representation is decoded and projected back to the pixel domain, recovering the encoded image.
The models suffer however a limited ability to adapt to images from a domain different than that of the training samples; 
standard video codecs include ad-hoc coding tools to deal with such contents~\cite{hevc}, as a testament to their relevance.
A straightforward approach to domain adaptation consists of fine-tuning the model over samples from the target domain.
However, that comes at the risk of \emph{catastrophic forgetting}, i.e. a performance loss on the source domain. 
In~\cite{tsubota2023} image-specific adapter modules are plugged into the encoder, however, this method requires training the adapters at encoding time and delivering their parameters to the decoder for each single image, jeopardizing its practical deployability. 
\\
In this work, we propose a novel method for LIC domain adaptation based on blending the contributions of domain-specific adapters plugged at the decoder side. 
Instead of training a separate adapter for each image~\cite{tsubota2023}, an adapter module is plugged into the decoder for both each target domain(s) and the source one. 
The adapters' contributions are blended according to the weights computed by a \emph{gate} network that infers the domain of the image to encode.
The adapters and the gate network are plugged into a pre-trained model that needs neither retraining nor refinement. 
The adapters and the gate are jointly trained and each adapter specializes in a specific domain,  whether it is one of the target domains or the source one.
\\
Our approach improves rate-distortion performance on the target domains, avoiding catastrophic forgetting on the source one. Furthermore, the adapters enhance image reconstruction even in classes that were never seen at training time.
Besides being effective, adapters do not modify parameters related to the original pre-trained model; this ensures that even if these components were not available during decoding, the image would be reconstructed without compromising the performance of the original pre-trained model.

\section{Background and related works}

This section provides some background on LIC and reviews the relevant literature regarding domain adaptation in such a context.

\subsection{Learned image compression} \label{sec_lic}
In LIC, a convolutional parametric autoencoder is trained end-to-end at compressing an image~\cite{review}.
In a nutshell, the encoder receives an image $\mathbf{x}$ as input and projects it to a latent representation $\mathbf{y}$ that is quantized into $\mathbf{\hat{y}}$ and then entropy coded, yielding a compressed representation in the form of a bitstream. At the receiver side, a decoder projects this representation back to the original dimension, recovering an approximation of the original image  $\mathbf{\hat{x}}$.
The entire model is trained by optimizing a Rate-Distorsion loss function $\lambda D + R$, where $D$ is the reconstruction error, $R$ is the rate term, and $\lambda$ is a hyperparameter that regulates the trade-off between the two.
In seminal works like~\cite{balle17}, the latent representation is modeled as a fully factorized distribution extracted either with an auxiliary neural network or in an analytical way~\cite{presta23}.
\cite{balle18} improved this approach by adding a \emph{scale hyperprior} that aims to find spatial correlation within an image, using Gaussian priors. 
In~\cite{minnen18}, a local context model based on mask convolution is used to enhance entropy estimation, while~\cite{minnen2020} channel-wise contexts are exploited to decrease computing time.
In the same wave, works like~\cite{cheng20,zou2022} make use of local and window-based attention, while~\cite{lu2022,liu23} exploited a mixture of CNN and transformer.
In general, the last models outperform conventional codecs on natural images; however, when transitioning to other specific domains, the emerging expected performance gap with traditional codecs narrows.

\subsection{Task-domain adaptation for learned image compression} \label{sec_ad}

To the best of our knowledge, task-domain adaptation for LIC has not been exhaustively studied.
In~\cite{danice21}, the parameters related to the generalized divisive normalization layer~\cite{gdn} and the entropy model are fine-tuned, with the addition of a small number of custom channel-specific parameters; despite this strategy, the authors were unable to eradicate the problem of forgetting.
Another possible approach to perform efficient domain adaptation is through the adapters~\cite{sung2022}, consisting of small modules added in the pre-trained model.
In that sense, we identify~\cite{tsubota2023} as the closest work to our method: Tsubota~\emph{et~al.} exploit tiny adapters introduced on the second attention module in the decoder to adapt the model to a single image. 
Specifically, for each image, they optimize the latent representation rate-distortion-wise, and then they tune the adapter parameters minimizing a loss function that combines distortion and parameter rate. Despite its effectiveness, this work has some drawbacks. First, the parameters of the adapters are part of the bitstream (since they are specific for each image), which causes transmission overhead. Second, Tsubota~\emph{et~al.} do not perform actual domain adaptation, but rather single-image content adaptation. Lastly, the encoding phase aligns with the optimization of the latent representation and adapter parameters, making the latter quite computationally expensive in terms of time and resources. On the other hand, we specialize adapters for multiple domains, either target or source, and we blend them to improve the final reconstruction, eliminating thus the need to encode adapter weights and simplifying the encoding phase, which remains unchanged compared to the model without adapters.

\section{Proposed method and architecture} \label{prop_method}

In this section, we describe our method and architecture for domain adaptation exemplifying the state-of-the-art Zou~\emph{et~al.}~\cite{zou2022} model.
Yet, our method is architecture-agnostic as we experimentally show later on over the Cheng~\emph{et~al.}~\cite{cheng20} model.
In a nutshell, we plug $K+1$ residual adapters into the decoder model pre-trained on a given source domain, i.e. $K$ adapters are meant for novel target domains and one is for the legacy source domain. 
The adapter on the source domain could be skipped but it turns out to be useful for fine-tuning. 
At decoding, we exploited such adapters along with a gate network $\varphi$ that yields a probability distribution over the $K+1$ domains $\mathbf{v}\!\in\![0,1]^{K+1}$, that is used to blend adapters' outcomes.
\\
The rest of this section is structured as follows. In Sec.~\ref{deconvad} and Sec.~\ref{gatenet} we first discuss the structure of the adapters and the gate, respectively. 
Next, in Sec.~\ref{merging}, we present the policy used to blend adapters' outputs. Finally, in Sec.~\ref{flow}, we describe the complete workflow of the resulting architecture along with the associated cost function and training process.

\subsection{Domain adaptation at the decoder}
\label{deconvad}

\begin{figure*}[!t]
    \begin{subfigure}{0.27\textwidth}
        \centering
        \includegraphics[width=\textwidth]{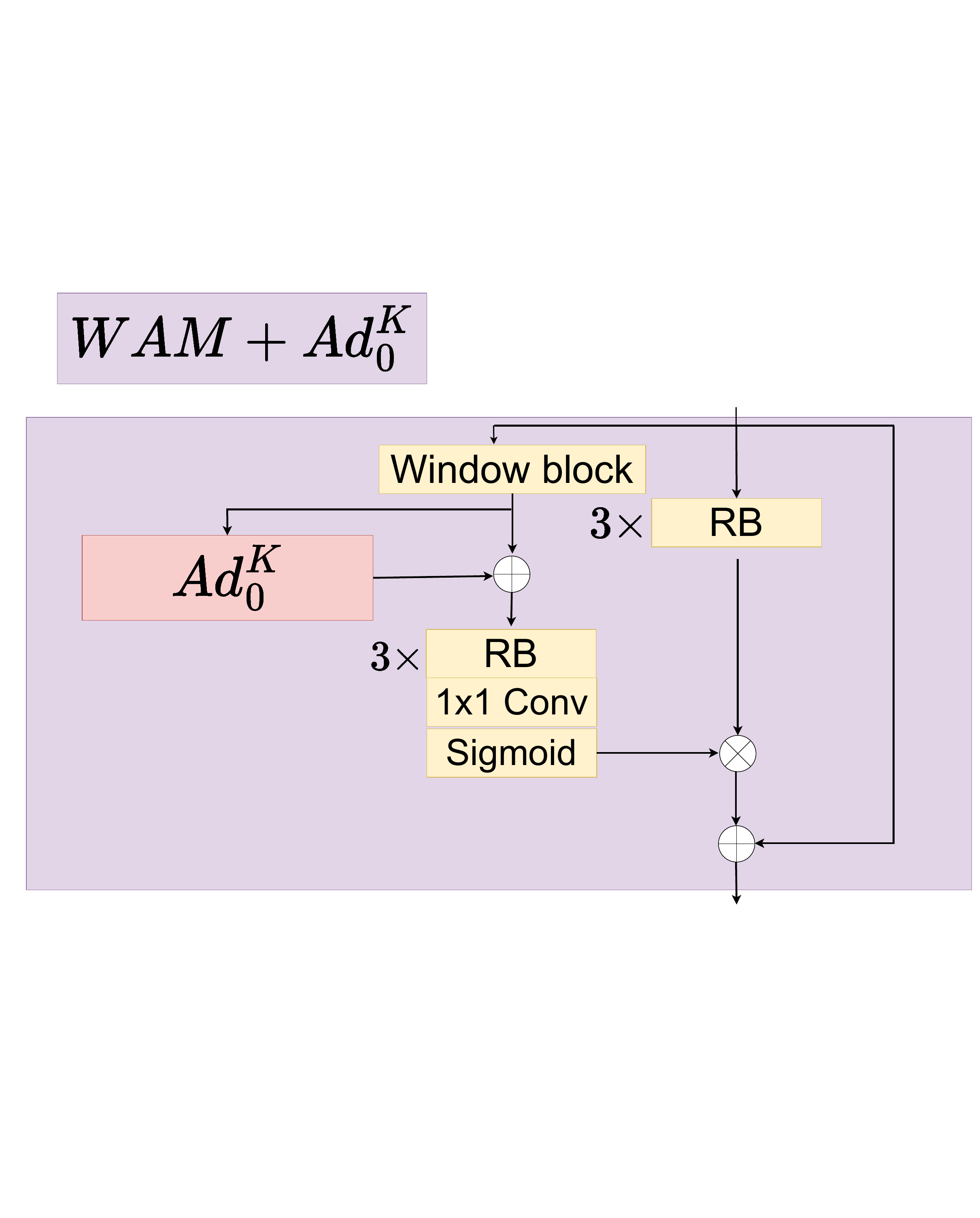}
        \caption{\emph{WAM adapters }}
        \label{wam_adapters}
  \end{subfigure} 
    \begin{subfigure}{0.27\textwidth}
        \centering
        \includegraphics[width=\textwidth]{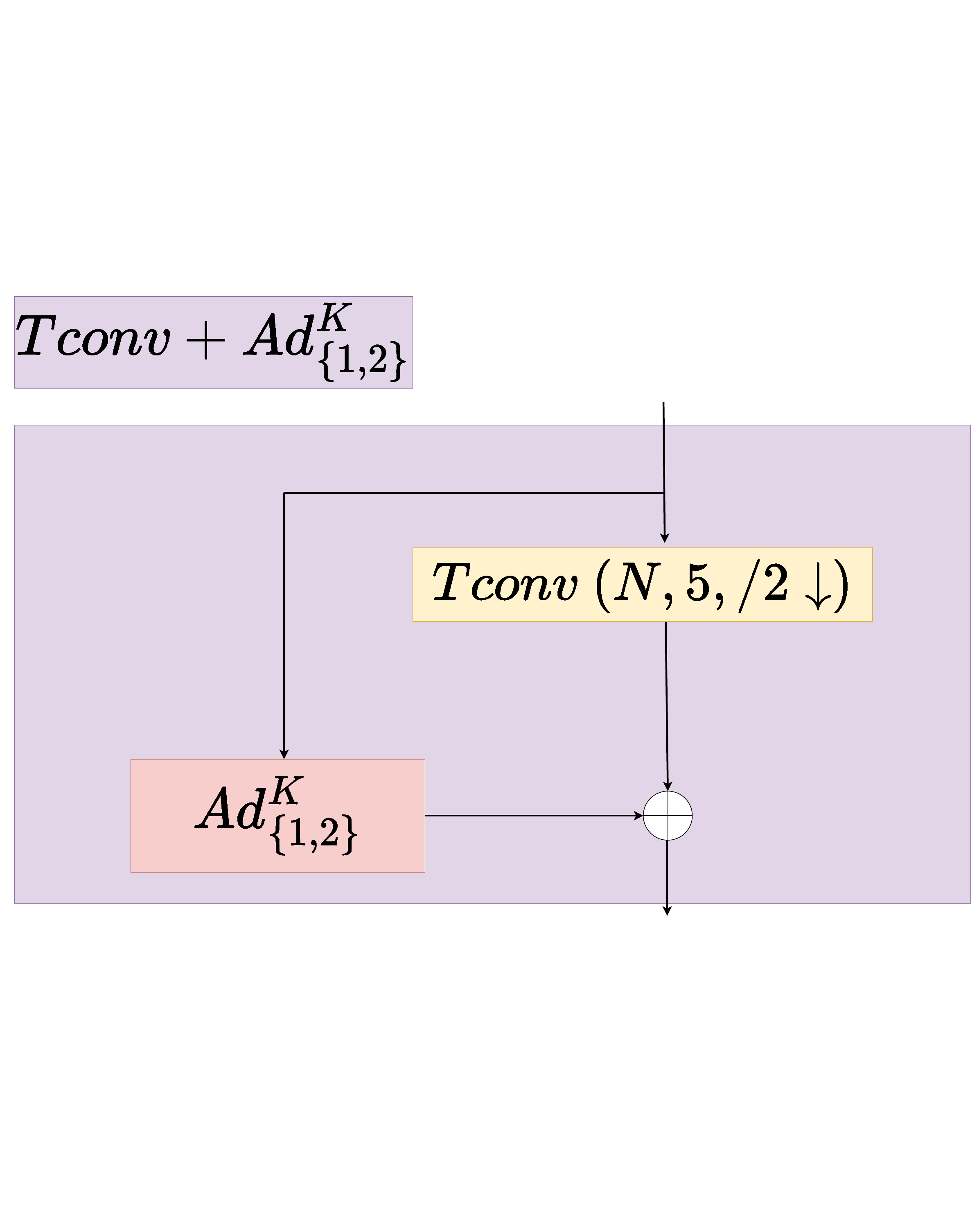}
        \caption{\emph{Tconv. adapters }}
        \label{deconv_adapters}
  \end{subfigure} 
     \begin{subfigure}{0.225\textwidth}
        \centering
        \includegraphics[width=\textwidth]{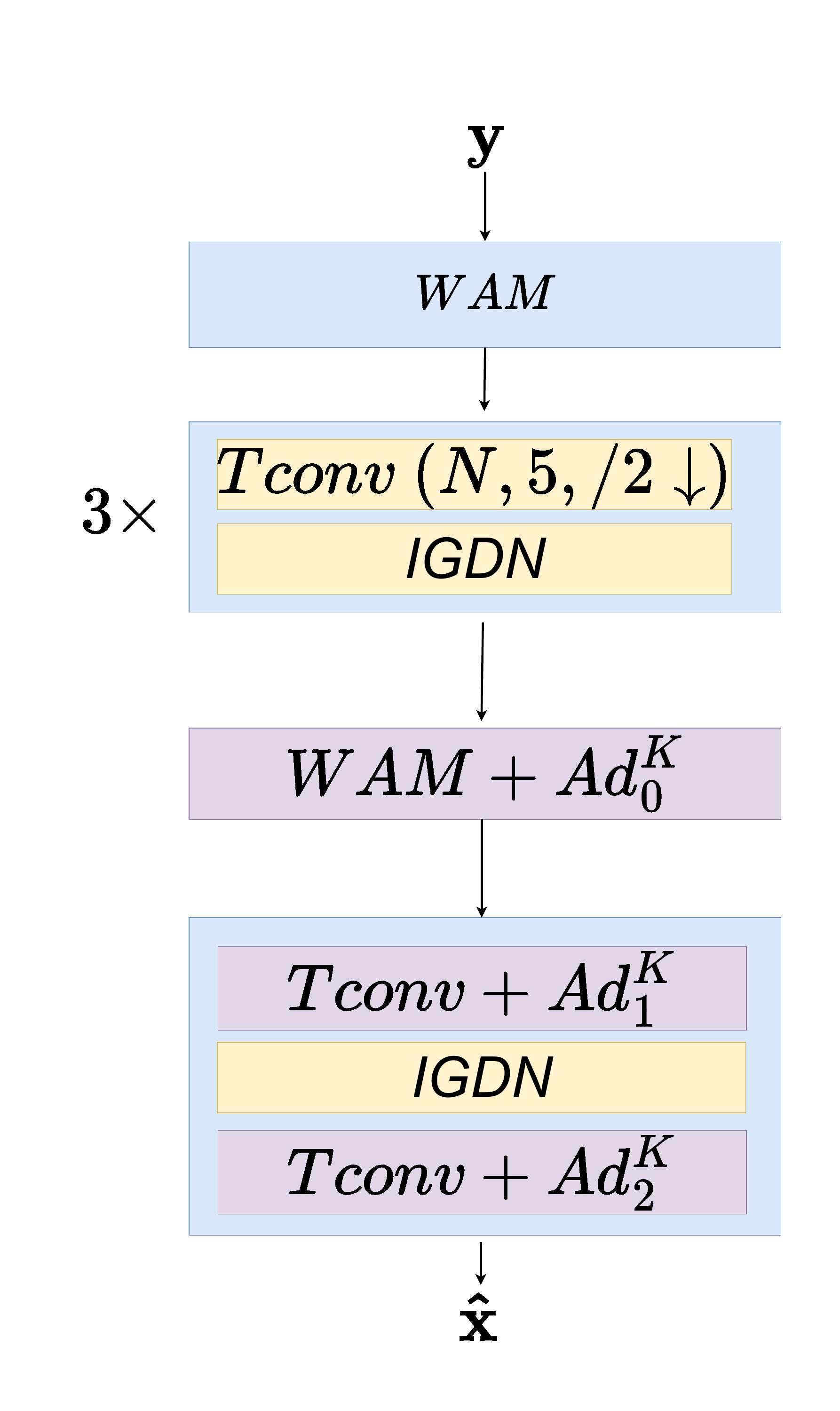}
        \caption{\emph{Decoder}}
        \label{dec}
  \end{subfigure}
    \begin{subfigure}{0.20\textwidth}
        \centering
        \includegraphics[width=\textwidth]{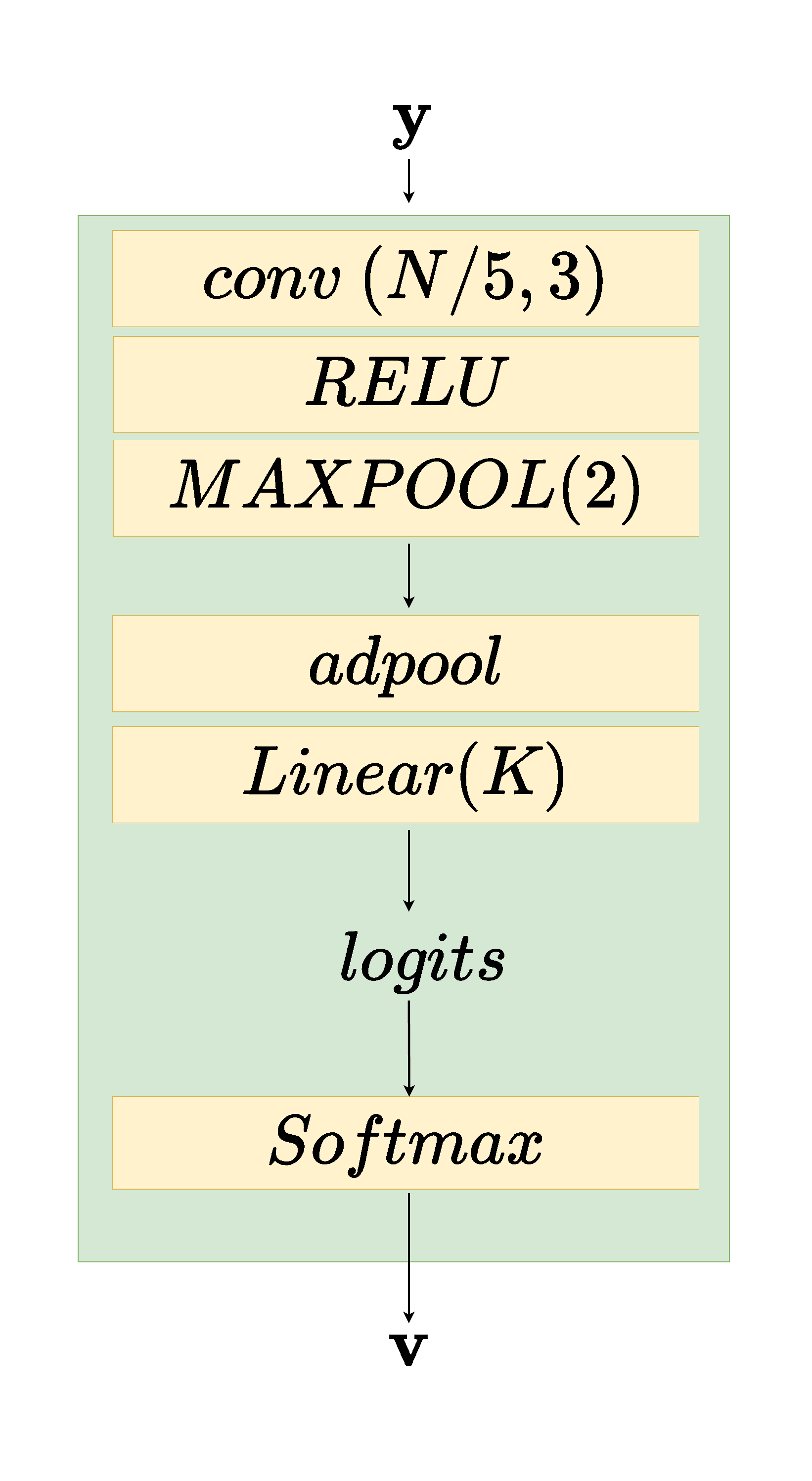}
        \caption{\emph{Gate network }}
        \label{agg_strategy}
  \end{subfigure} 
  \caption{(a) Architecture of the WAM adapter $Ad^{k}_{0}$. (b) Architecture of the  deconvolutional adapters $Ad^{k}_{\{1,2\}}$. (c) Overall decoder. (d) Gate network $\varphi$.}
  \label{adapters_fig}
\end{figure*}

In our architecture, one adapter $\mathbf{Ad}_{:}^{k}$  is plugged into the decoder for each of the $K+1$ domains of interest, i.e. $K$ target domains plus the source one; 
Each adapter is composed by three modules $\mathbf{Ad}_{:}^{k} = \{Ad^{k}_{0}, Ad^{k}_{1}, Ad^{k}_{2}\}$, each one including one convolutional layer with $3 \times 3$ kernels.
As shown in Fig.~\ref{adapters_fig}(a), the module $Ad^{k}_{0}$ is plugged into the second Window Attention Module (WAM) block of the decoder, as in~\cite{tsubota2023}, to enhance the attention towards the more detailed parts of the image. 
Namely, the convolutional layer receives in input the output of the WAM block, preserving the number of channels and spatial dimensions.
The remaining $Ad^{k}_{1}$ and $Ad^{k}_{2}$ modules are plugged at the output of the last two transpose-convolutional layers respectively, as in Fig.~\ref{adapters_fig}(b).
For the sake of similarity, these two modules have the same structure as the respective layers they are plugged into, consisting of a deconvolutional layer that doubles the spatial dimensionality of the input feature maps.
With respect to comparable literature where adapter modules are plugged into the encoder, our choice of plugging adapters into the decoder bears some advantages.
First, the bitstream produced by the pre-trained reference encoder can be reused as it is.
Second, training the adapters at the decoder avoids the cost of backpropagating the gradients to the encoder.
The overall decoder is shown in Fig.~\ref{adapters_fig}(c).

\subsection{Gate Network} \label{gatenet}

The Gate network $\varphi$ takes as input the latent representation $\mathbf{y}$ and outputs a domain probability distribution.
The gate architecture is shown in Fig.~\ref{adapters_fig}(d) and includes one convolutional layer, a ReLU activation, a MaxPooling layer for complexity control, and an adaptive pooling layer that allows handling inputs of arbitrary size.
A linear layer then yields $K+1$ classification logits that are normalized into the probability distribution $\mathbf{v} = (v_{0},..,v_{K})$ by a Softmax layer and delivered to the decoder.
Such distribution amounts to $K+1$ real-valued coefficients that are compressed and embedded in the bitstream beside the compressed latent representation $\mathbf{\hat{y}}$.
While the compression of such coefficients is out of the scope of this work, their rate can be safely assumed negligible to the compressed latent representation.
For example, for a sample image from the classic Kodak dataset, the weight to encode $\mathbf{v}$ is on the order of $10^{-4}$ compared to the compressed latent representation.
The gate could be in principle plugged into the decoder, sampling the compressed latent representation $\mathbf{\hat{y}}$. However, in this work we refrain our experiments from sampling the uncompressed latent representation $\mathbf{y}$, leaving the effect of quantizing the latent representation for our future endeavors.
Furthermore, using $\mathbf{y}$ as the input instead of the original image allows us to leverage the latent representation already extracted by the encoder, thus simplifying the gate's task.

\begin{figure*}[!t]

  \hfill 
    \begin{subfigure}{0.32\textwidth}
        \centering
        \includegraphics[ width=\textwidth]{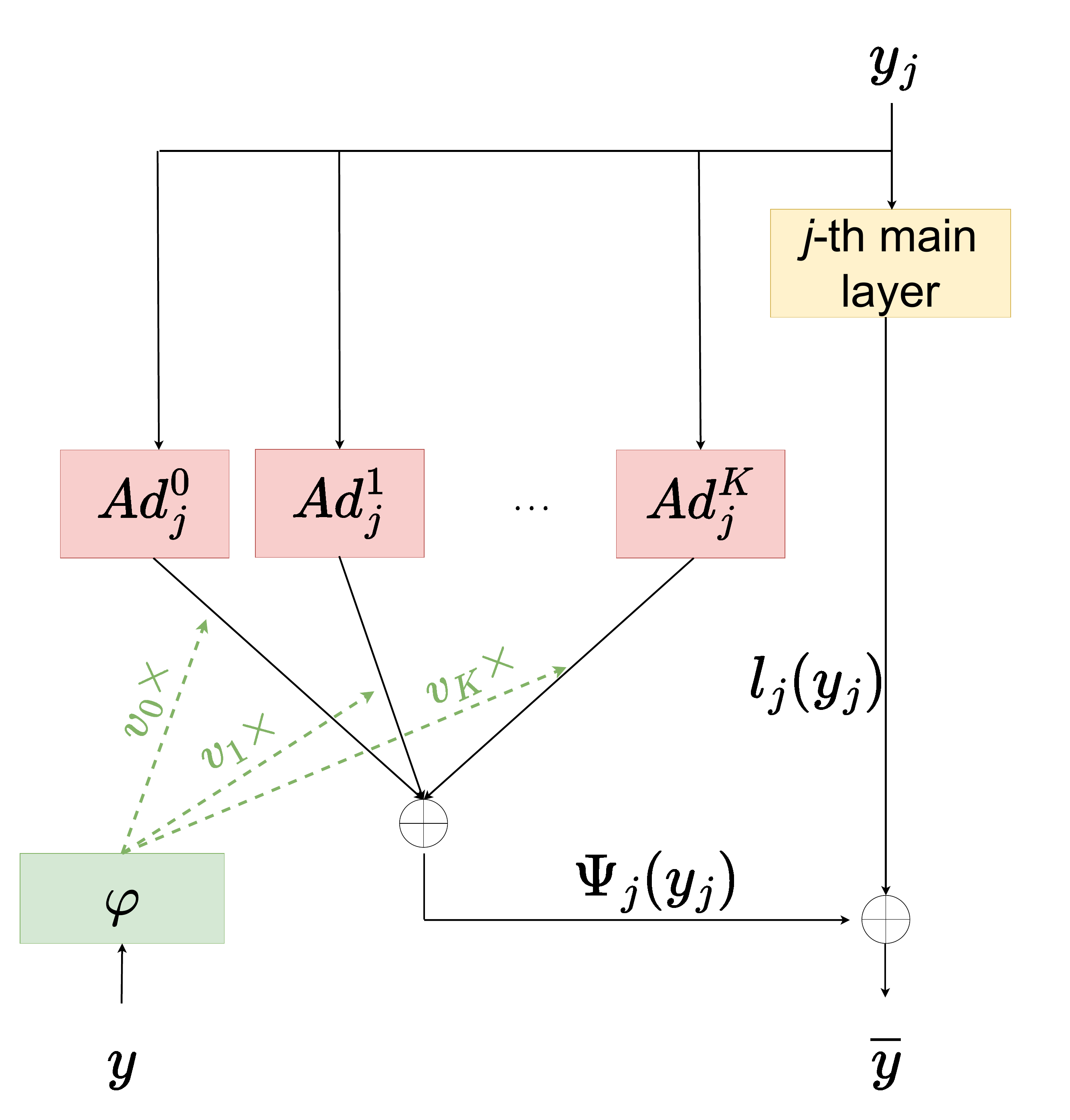}
        \caption{\emph{blending policy.} }
        \label{gate_gen_b}
  \end{subfigure} 
  \hfill
      \begin{subfigure}{0.55\textwidth}
        \centering
        \includegraphics[width=\textwidth]{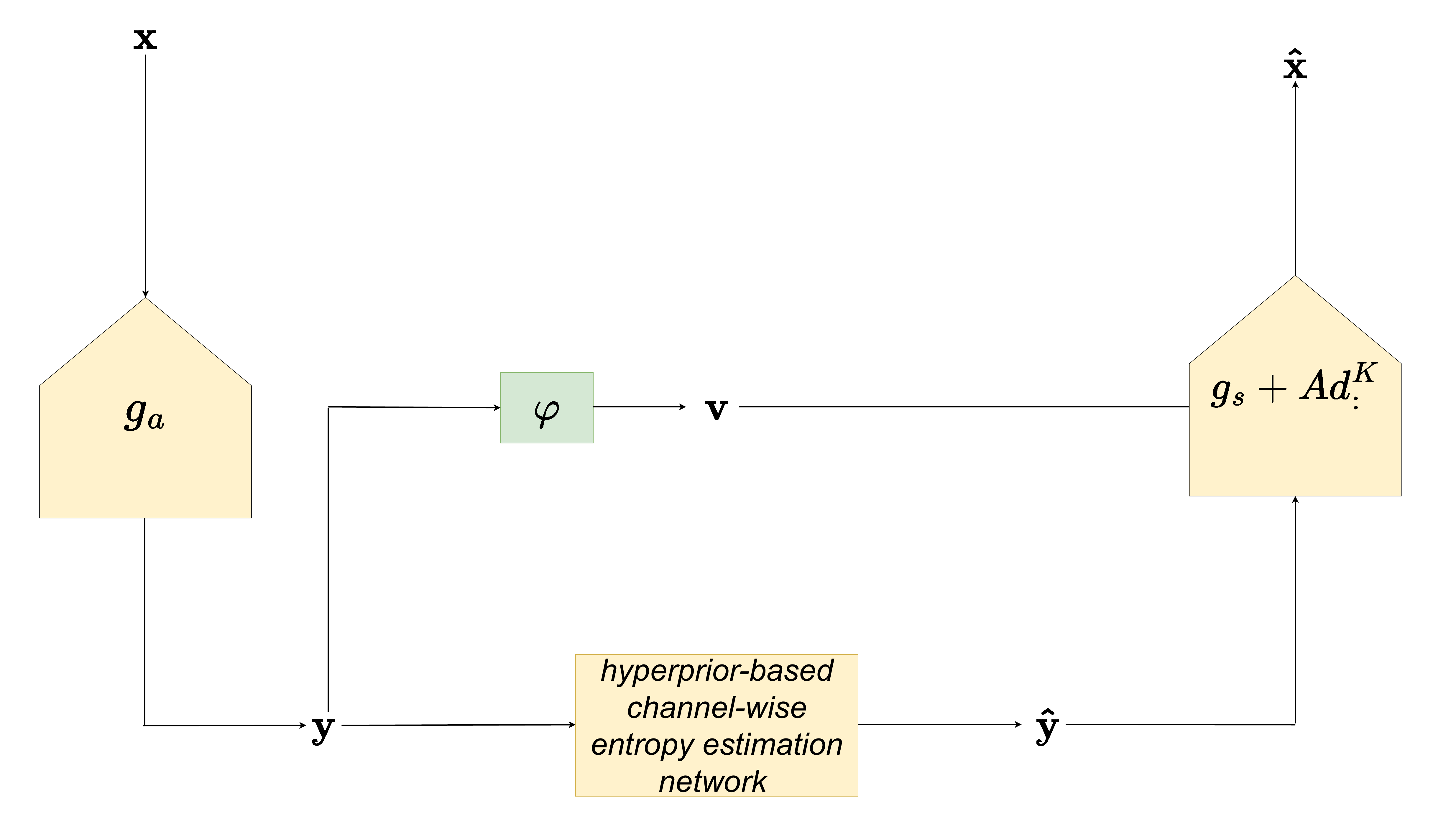}
        \caption{\emph{Overall LIC model architecture with $\varphi$.}}
        \label{gate_gen_a}
  \end{subfigure}
  \caption{(a) Adapters blending policy based on weighted sum. Index 0 refers to the source domain adapter.  (b) Blueprint of a LIC model with the gate.}
  \label{gate_gen}

\end{figure*}

 \subsection{Adapters blending policy } \label{merging}

The outputs of the adapters are blended at the decoder according to the distribution coefficients $\mathbf{v}$ embedded in the bitstream. 
Namely, the probability distribution $~{\mathbf{v}\!=\!(v_{0},..,v_{K})}$ among domains of interest, i.e. target and source, is used to compute a weighted average between the adapters outputs. 
Let $\mathbf{y}_{j}$ be the input of the $j$-th layer of the decoder with the adapter ($j = \{0,1,2\}$), adapters-enhanced output $\mathbf{\overline{y}}$ is calculated as

\begin{equation}
   \mathbf{\overline{y}} = \Psi_{j}(y_j) + l_{j}(y_j),
   \label{sum_of_adapter}
\end{equation} 
\noindent
where $l_{j}$ represents the $j$-th layer of the decoder, and the addition operator is element-wise sum. Then we have

\begin{equation}
    \Psi_{j}(y_j) = \sum_{k = 0}^{K} v_{k} \cdot Ad_{j}^{k}(y_j),
    \label{adapter_weighted_sum}
\end{equation} 
\noindent
where $Ad_{j}^{k}$ is the $j$-th level adapter of the $k$-th domain.
The blending policy just described is shown in Fig.~\ref{gate_gen}(a), where we refer to the pre-trained layer network without adapter, as the main layer.

\subsection{Training the adapters and the gate} \label{flow}

This section first describes the workflow of our method and then how both adapters and the gate are trained given a model pre-trained on the source domain, e.g. natural images.
The training set $X_t$ is composed of $T$ images uniformly distributed among the $K$ target domains with the corresponding domain labels $d_t$, with $t\!=\!1,...,T$.
As shown in Fig.~\ref{gate_gen}(b), an image $\mathbf{x}_{t}$ is passed to the encoder $g_{a}$ producing the latent representation $\mathbf{y}$, which is first fed as input to the \emph{gate network} $\varphi$ that yields a probability distribution over the $K+1$ domains $\mathbf{v}\!\in\![0,1]^{K+1}$. 
The probability vector $\mathbf{v}$ is then sent to the decoder beside the quantized latent representation $\mathbf{\hat{y}}$, which is encoded through the channel-wise entropy estimation module~\cite{minnen2020}, and then it is used to weight the output of each adapter. 
Once decoded back,  $\mathbf{\hat{y}}$ is passed to our adapter-based decoder $g_{s} + \mathbf{Ad}_{:}^{K}$ , returning the reconstructed image $\mathbf{\hat{x}_{t}}$.
\\
Once initialized from a Gaussian distribution, both the adapters $\mathbf{Ad}_{:}^{k}$ and the gate $\varphi$ are jointly trained for all domains of interest at once, freezing the other parameters of the model.
The cost function minimized at training time is the sum between the reconstruction error $D$ and a \emph{domain mismatch term}, that should force $\varphi$ to learn different domains from $\mathbf{y}$.
By casting this as a classification problem, the objective is to minimize the cross-entropy ($\ce$) between the distribution $\mathbf{v}$ derived from $\varphi$ and the true labels $d_{t}$.
Summing up, the minimized cost function is

\begin{equation}\label{distorsion_loss}
    \mathcal{L} = \gamma \cdot \mse(\mathbf{x}_{t}, \mathbf{\hat{x}}_{t}) + \ce(d_{t}, \mathbf{v})
\end{equation}
\noindent
where $\gamma$ is a hyperparameter that regulates the trade-off between these two terms, and $\mse$ is the \emph{mean squared error} minimizing the distortion.
Notice that the second term in~\eqref{distorsion_loss}
does not  represents the rate of $\mathbf{\hat{y}}$, typically minimized in the traditional RD loss function. 
Since we do not refine the encoder, the compressed latent representation $\mathbf{\hat{y}}$ remains untouched as well as its rate, allowing us to drop the rate component $R$ from the minimized cost function.

\section{Experiments and Results}

This section presents the results obtained with our method over the Zou~\emph{et~al.}~\cite{zou2022} and Cheng~\emph{et~al.}~\cite{cheng20} models; concerning the latter, we added the deconvolutional adapters in the last three layers of the decoder, used in the same way as for Zou~\emph{et~al.}~\cite{zou2022}.
We experiment with adapting the pretrained models over two different target domains gauging the performance on the target as well as the source domain.
Our implementation\footnote{ our code is available online at \url{https://github.com/EIDOSLAB/LIC-Domain-Adaptation-with-supervised-Adapters}.} leverages on the CompressAI library~\cite{compressai}.

\subsection{Experimental setup}\label{setup}

We considered a total of $K\!=\!2$ target domains, namely the \textit{Sketch} and \textit{Comic}  which are added to the source domain used for pretraining the model, e.g. natural images.
Towards training the gate and the adapters, we had to combine different sources due to the lack of high-resolution datasets for domain adaptation in LIC. 
For the natural domain, we used the OpenImages dataset~\cite{openimages}; for the sketch domain, we used  ImageNet-Sketch~\cite{wang2019}; for the comic domain, we used BAM~\cite{bam}.
Namely, we randomly sampled 4k images from each dataset, resulting in a total of $T\!=\!12$k training images.
During training, we froze all the parameters but those related to the adapters and the gate, i.e. the pre-trained encoder and decoder are not refined. 
We trained gate and adapters for $400$ epochs using Adam with an initial learning rate of $10^{-4}$, halving it when reaching a plateau with the patience of 15 epochs and batch size of $16$ and with $\gamma = 0.5$ in \eqref{distorsion_loss}.
At inference time, we assess the performance of our method over four distinct test datasets.
The target sketch and comic domains are represented by 100 random images from ImageNet-Sketch and BAM that are not present in the training set.
The source domain is represented by the Kodak~\cite{kodak} and CLIC~\cite{clic}.
The performance of our method is reported in terms of RD curves, BD-Rate, and BD-PSNR.
Concerning complexity, the adapter modules include about \num{3.8}M
parameters, i.e. $\sim 5.5$\% of Zou~\emph{et~al.} decoder complexity.
Similarly, the gate is composed of \num{184}k 
parameters, i.e. $\sim0.28$\% of Zou~\emph{et~al.} encoder complexity.

\begin{figure*}[!t]

    \begin{subfigure}{0.400\textwidth}
        \centering
        \includegraphics[width=\textwidth]{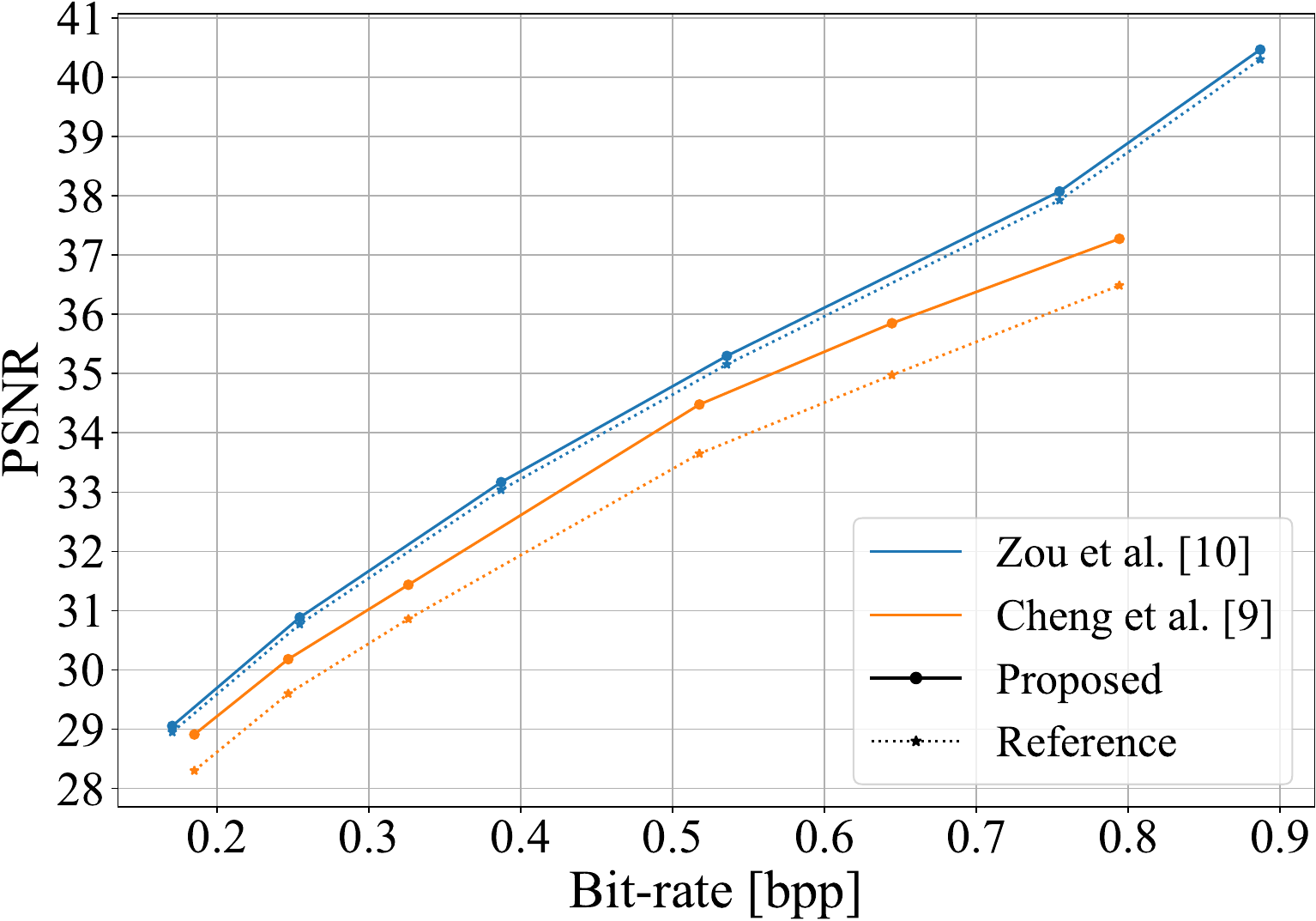}
        \caption{\emph{Sketch}}
        \label{sketch}
  \end{subfigure} 
\hfill
    \begin{subfigure}{0.400\textwidth}
        \centering
        \includegraphics[width=\textwidth]{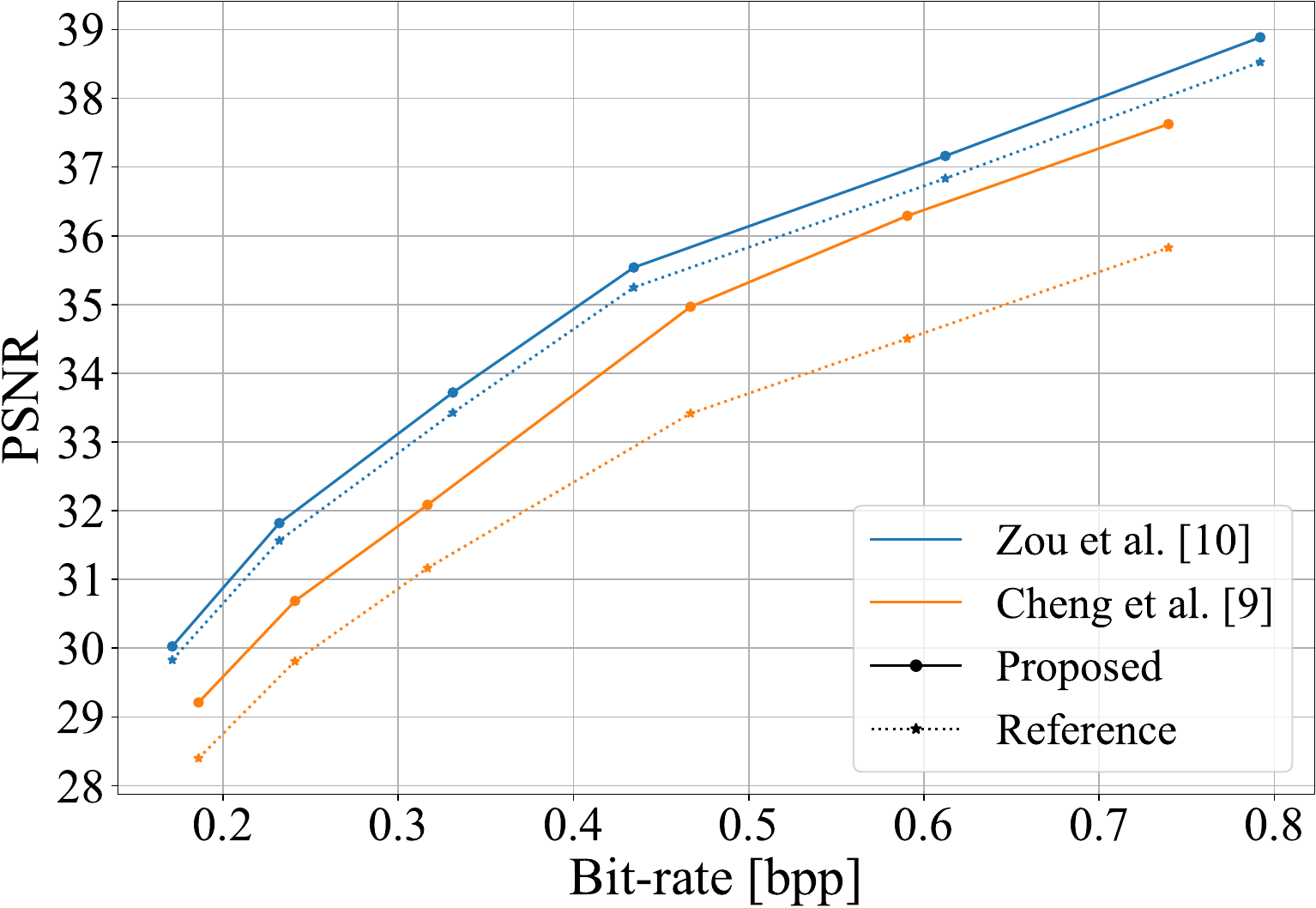}
        \caption{\emph{Comic}}
        \label{bam_comic}
  \end{subfigure} 
  \caption{Rate-PSNR plots for our Adapter-based method vs Reference for Zou~\emph{et~al.} (blue) and Cheng~\emph{et~al.} (orange), considering Sketch (a) and Comic (b).}
  \label{psnr_res}
\end{figure*}

\begin{figure}
  \ffigbox{
    \begin{subfloatrow}
      \ffigbox{\caption{BD-Rate/PSNR for different domains.}\label{tab_bdrate}}{
        \resizebox{0.55000\textwidth}{!}{
        \begin{tabular}{l   l l l   l l    }
        \toprule
         \multicolumn{1}{c}{\textbf{\emph{Dataset}}}  & 
         \multicolumn{1}{c}{\textbf{\emph{Domain}}} &
         \multicolumn{2}{c}{\textbf{\emph{Zou~\emph{et~al.}}}}  & \multicolumn{2}{c}{\textbf{\emph{Cheng~\emph{et~al.}}}}  \\
        \cmidrule(l){1-1}\cmidrule(l){2-2}\cmidrule(l){3-4} \cmidrule(l){5-6} 
         & & \multicolumn{1}{c}{BD-Rate}   &   \multicolumn{1}{c}{BD-PSNR}   &  \multicolumn{1}{c}{BD-Rate}   & \multicolumn{1}{c}{BD-PSNR}   \\
        \midrule
        \emph{Kodak} & \emph{Natural} &\multicolumn{1}{c}{0.0012}  & \multicolumn{1}{c}{ $\sim 0$}  & \multicolumn{1}{c}{-1.46} &\multicolumn{1}{c}{0.06}  \\
        \emph{clic}  & \emph{Natural} & \multicolumn{1}{c}{0.038}  & \multicolumn{1}{c}{$\sim 0$}  & \multicolumn{1}{c}{-1.75} & \multicolumn{1}{c}{0.08}   \\
        \emph{Sketch}  & \emph{Sketch} &\multicolumn{1}{c}{-2.45}  & \multicolumn{1}{c}{0.1718}  & \multicolumn{1}{c}{-11.55} & \multicolumn{1}{c}{0.69}   \\
        \emph{Comic}  & \emph{Comic} & \multicolumn{1}{c}{-4.93 }  & \multicolumn{1}{c}{0.28}  & \multicolumn{1}{c}{-19.15} & \multicolumn{1}{c}{1.25}   \\
        
        \bottomrule
        \end{tabular}
        }
       
      }
      \hfill
      \ffigbox{\caption{Avg. probabilities predicted by $\varphi$.} \label{vp}}{\includegraphics[width=0.230\textwidth]{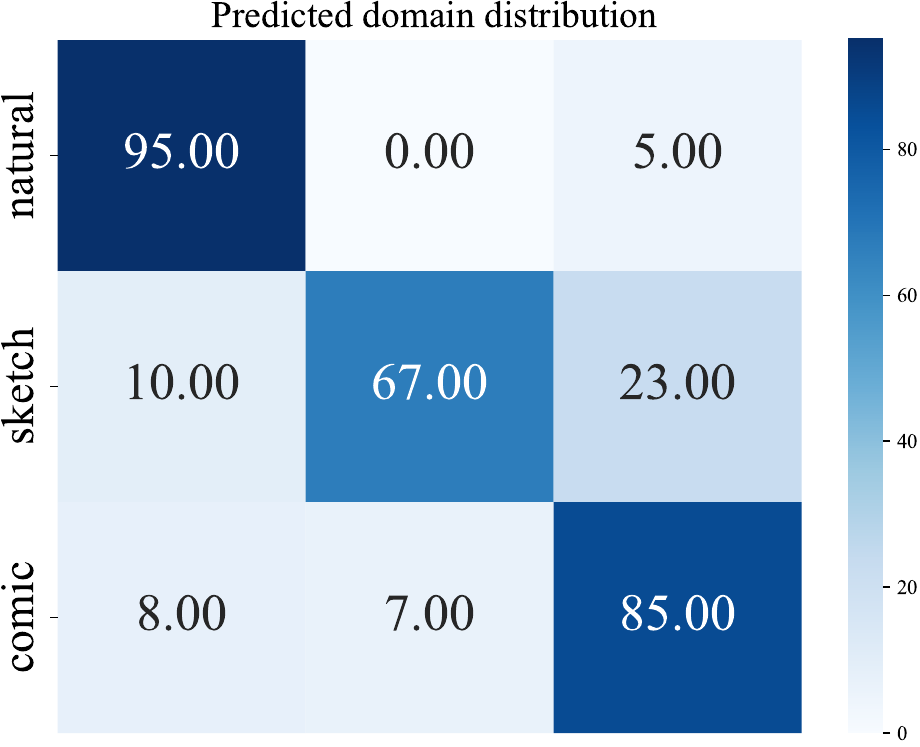}}
    \end{subfloatrow}
  }
  {\caption{(a) BD-Rate/PSNR for different domains, considering both Zou~\emph{et~al.} and Cheng~\emph{et~al.} as anchor, with our models as reference. (b) Avg. probabilities $\mathbf{v}$ predicted by the Gate $\varphi$, starting from Zou~\emph{et~al.} for the second-best quality model.}}
\end{figure}

\subsection{Rate-Distortion Performance}\label{overall_perf}

\begin{figure*}[!h]

    \begin{subfigure}{0.29\textwidth}
        \centering
        \includegraphics[width=\textwidth]{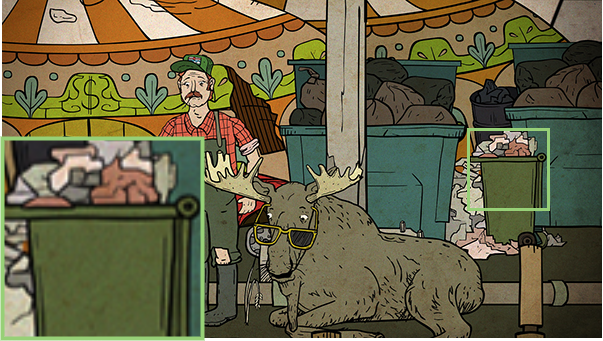}
        \caption{Original Image}
        \label{orig}
  \end{subfigure} 
  \hfill 
    \begin{subfigure}{0.290\textwidth}
        \centering
        \includegraphics[width=\textwidth]{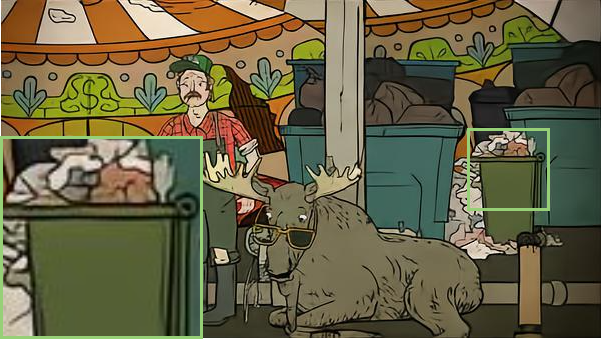}
        \caption{Zou~\emph{et~al.} PSNR: 25.62 }
        \label{imgred}
  \end{subfigure} 
    \hfill 
    \begin{subfigure}{0.29\textwidth}
        \centering
        \includegraphics[width=\textwidth]{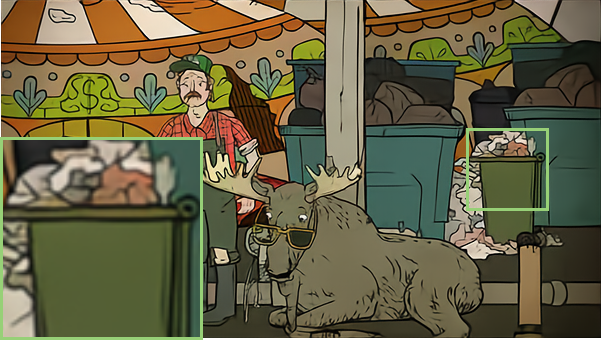}
        \caption{Prop. PSNR: 25.90}
        \label{img_prop}
  \end{subfigure} 
  \caption{Reconstruction of a sample taken from BAM~\cite{bam} (from Comic domain). (a) Original, (b) Zou~\emph{et~al.}, (c) Proposed. Both (c) and (d) have bpp equal to 0.31.}
  \label{rec_sample}
\end{figure*}

Fig.~\ref{psnr_res} shows the RD performance of our method for the Zou~\emph{et~al.} and Cheng~\emph{et~al.} models over the target sketch and comic domains. 
Our method visibly improves the performance for both the target domains to the pre-trained model. 
Table ~\ref{tab_bdrate} summarizes the two models' performance in terms of BD-Rate and BD-PSNR gains over the four test datasets.
Our method strikes rate reductions of 5\% and 10\% over the comic and sketch domains.
Yet, we achieve some gain also over the source domain(Kodak and Clic datasets) for Cheng~\emph{et~al.}, proving our domain adaptation method does not incur in any \emph{catastrophic forgetting}.
The BD-Rate reduction is remarkable Especially for Cheng~\emph{et~al.}: we hypothesize this model is less able to generalize outside the source domain than Zou~\emph{et~al.}, hence our measured gains.
Fig.~\ref{vp} shows how the gate network blends the outputs of the adapters across different domains for Zou~\emph{et~al.}.
For all domains the adopted training policy induces the exploitation of all the available adapters to increase the performance since; even though the gate is capable of identifying the predominant domain for an image, it still utilizes the others to a lesser extent.
Fig.~\ref{rec_sample} shows samples of decoded images: our method preserves best the fine-grained details over the target domains.


\begin{table*}[!h]
\centering
\resizebox{0.5\textwidth}{!}{
\begin{tabular}{ccccc  }
\toprule
 \multicolumn{1}{c}{\textbf{\emph{Dataset}}}  & \multicolumn{1}{c}{\textbf{\emph{Reference}}}  & \multicolumn{3}{c}{\textbf{\emph{Blending method}}}  \\
\cmidrule(l){1-1}\cmidrule(l){2-2} \cmidrule(l){3-5} 
 &   &     \multicolumn{1}{c}{Proposed}   & \multicolumn{1}{c}{Top1}  & \multicolumn{1}{c}{Oracle}      \\
\midrule
\emph{Kodak (Natural)} & \multicolumn{1}{c}{35.947}  & 
\multicolumn{1}{c}{35.945}  & \multicolumn{1}{c}{35.948} &\multicolumn{1}{c}{\textbf{\textbf{35.950}}}   \\
\emph{Clic (Natural)}  & \multicolumn{1}{c}{36.618}  & \multicolumn{1}{c}{36.613}  & \multicolumn{1}{c}{36.614} & \multicolumn{1}{c}{\textbf{36.618}}   \\
\emph{Sketch} &  \multicolumn{1}{c}{37.925}   & \multicolumn{1}{c}{\textbf{38.112}}    &\multicolumn{1}{c}{38.005} &\multicolumn{1}{c}{38.109}  \\
\emph{Comic} &  \multicolumn{1}{c}{36.833}   & \multicolumn{1}{c}{\textbf{37.161}}    &\multicolumn{1}{c}{37.138} &\multicolumn{1}{c}{37.141}   \\

\bottomrule
\end{tabular}
\vspace{-7pt}
}
\caption{PSNR on different test sets considering different blending policies.}\label{tab-psnr}
\end{table*}

\subsection{Comparison with other blending policies} \label{aggregation_policy}

Next, we assess the impact of the policy used to blend the adapters modules outputs.
We recall that with our \textit{proposed} method, the outputs of the adapters are weighted according to the probability distribution predicted by the gate network $\varphi$. 
We now consider two other blending approaches, namely \emph{top1} and \emph{oracle}.
The former considers only the output of the adapter corresponding to the top-scoring class predicted by the gate, dropping the contributions from other adapters.
The latter is a hypothetical scheme that relies on the assumption that the image class label is known at testing time, discarding any input from the gate.
We highlight that these policies are used not only at inference but also during the training phase.
Tab.~\ref{tab-psnr} shows the results in terms of PSNR considering the second highest quality model in Fig.~\ref{psnr_res}.
The \emph{proposed} policy outperforms the other two for both target domains, meaning that exploiting all the adapters with the right blending maximizes the image quality.
On the other hand, the oracle policy yields the best image quality over natural images, i.e. the source domain.
However, the oracle policy relies on the hypothesis that the image domain is known for each image to encode, which is unrealistic in most cases.

\vspace{-5pt}
\subsection{Unseen domains}\label{unseen}
\vspace{-5pt}
We evaluate our methods for domains unseen both when pretraining the Zou~\emph{et~al.} model and at the time of training the adapters and the gate over the two target domains.
We considered different types of images from different datasets like BAM~\cite{bam}, DomainNet~\cite{domainnet}, and IIIT-AR-13K~\cite{documents}.
Tab.~\ref{tab_unseen} shows that for every domain our method yields better encoding efficiency, even if sometimes minimally.
However, for some domains the gains are remarkable; for example, the 20\% gain for \emph{Quickdraw} is likely to be attributed to its semantic similarity to the \emph{Sketch} domain.
This hypothesis is supported by the probability distribution generated by the gate network $\varphi$, where approximately 90\% of the image likelihood is assigned to the adapter associated with the Sketch domain.
For its possible practical application, an important result is the BD-Rate equal to $-2.45$ on~\cite{documents}, representing graphical and documents images: in this case, we see how $\varphi$ associated this domain more with the Sketch class, which is intuitive concerning the contents of this dataset.
In general, The addition of domain-specific adapters also brings further performance improvement, and therefore knowledge, in other typologies of images, thanks to increased generalization capacity and the capacity of the gate to spot correlations with known domains.

\begin{table*}[!h]
\centering
\resizebox{0.555\textwidth}{!}{
\begin{tabular}{l   l l   l l l   }
\toprule
 \multicolumn{1}{c}{\textbf{\emph{Dataset}}}  & \multicolumn{2}{c}{\textbf{\emph{Bjontegaard Metrics}}}  & \multicolumn{3}{c}{\textbf{\emph{ Predicted domain distribution}}}  \\
\cmidrule(l){1-1}\cmidrule(l){2-3} \cmidrule(l){4-6} 
 &  \multicolumn{1}{c}{BD-Rate}   &   \multicolumn{1}{c}{BD-PSNR}   &  \multicolumn{1}{c}{Natural}   & \multicolumn{1}{c}{sketch}  & \multicolumn{1}{c}{comic}      \\
\midrule
\emph{Infograph}~\cite{domainnet} & \multicolumn{1}{c}{-0.44}  & \multicolumn{1}{c}{0.007}  & \multicolumn{1}{c}{48} &\multicolumn{1}{c}{3} & \multicolumn{1}{c}{49}   \\
\emph{Bam-Drawing}~\cite{bam}  & \multicolumn{1}{c}{\textbf{-2.789}}  & \multicolumn{1}{c}{0.216}  & \multicolumn{1}{c}{12} & \multicolumn{1}{c}{42} & \multicolumn{1}{c}{46}  \\
\emph{Quickdraw}~\cite{domainnet} & \multicolumn{1}{c}{\textbf{-21.574}}  & \multicolumn{1}{c}{1.95}    & \multicolumn{1}{c}{0} & \multicolumn{1}{c}{93} & \multicolumn{1}{c}{7}    \\
\emph{watercolor}~\cite{bam} & \multicolumn{1}{c}{-0.61} &\multicolumn{1}{c}{0.02} & \multicolumn{1}{c}{49} & \multicolumn{1}{c}{4} & \multicolumn{1}{c}{17} \\
\emph{clipart}~\cite{domainnet} &  \multicolumn{1}{c}{\textbf{-1.32}} & \multicolumn{1}{c}{0.065}  & \multicolumn{1}{c}{15} & \multicolumn{1}{c}{0.07} & \multicolumn{1}{c}{78} \\
\emph{IIIT-AR-13K}~\cite{documents}  &  \multicolumn{1}{c}{\textbf{-2.45}} & \multicolumn{1}{c}{0.20}  & \multicolumn{1}{c}{15} & \multicolumn{1}{c}{50} & \multicolumn{1}{c}{35} \\
\bottomrule
\end{tabular} 
}
\vspace{-8pt}
\caption{\emph{First two columns}: Bjontegaard Metrics for unseen domains, considering \cite{zou2022} as reference.  \emph{Second three columns}: average $\mathbf{v}$ considering highest quality model.} \label{tab_unseen}
\end{table*}

\vspace{-5pt}
\section{Conclusions and future works}
\vspace{-10pt}

This study addressed the problem of domain adaptation in learned image compression.
Our method outperforms reference pre-trained state-of-the-art models on the target domains without forgetting the source one (Sec.~\ref{overall_perf}), showing also (Sec.~\ref{aggregation_policy}) that aggregating the adapters brings a slight performance improvement. 
In Sec.~\ref{unseen} we showed that our method offers advantages also over unseen domains.
This method could benefit from improvements like unsupervised gate training to remove the need for predefined adapter classes.
Additionally, domain adaptation at the encoding stage could enhance performance by modifying the entropy estimation.
\vspace{-10pt}
\section*{Acknowledgements}
\vspace{-10pt}
This work was partially funded by the Hi!PARIS Center on Data Analytics and Artificial Intelligence.

\vspace{-15pt}
\Section{References}


\end{document}